# Construction Inspection through Spatial Database

Ahmad Hasan[a], Ashraf Qadir[a], Ian Nordeng[a] and Jeremiah Neubert[a]

[a]Department of Mechanical Engineering, University of North Dakota, United States of America
E-mail: ahmadjarjis.hasan@und.edu, ashraf.qadir@und.edu, ian.nordeng@und.edu, jeremiah.neubert@und.edu

**Abstract –**
   This paper presents a novel pipeline for development of an efficient set of tools for extracting information from the video of a structure, captured by an Unmanned Aircraft System (UAS) to produce as-built documentation to aid inspection of large multi-storied building during construction. Our system uses the output from a Simultaneous Localization and Mapping system and a 3D CAD model of the structure in order to construct a spatial database to store images into the 3D CAD model space. This allows the user to perform a spatial query for images through spatial indexing into the 3D CAD model space. The image returned by the spatial query is used to extract metric information. The spatial database is also used to generate a 3D textured model which provides a visual as-built documentation.

## 1   Introduction

   The objective of our system is to reduce the amount of labor and time required to collect important construction site information, and enable engineers to generate as-built documentation of building elements. Generation of as-built documentation requires engineers to search through large amounts of video of the construction site for a specific region of interest. This is time-consuming and an inefficient method. The proposed system provides a credible solution to that problem by aligning a spatial database to the 3D CAD model in order to facilitate a spatial query for the images desired. This is done by allowing the user to click on the region of interest on the 3D CAD model.

   The system takes a previously generated 3D CAD model and the images collected with the UAS, which may be processed by a variety of keyframe based Simultaneous Localization and Mapping (SLAM) [1,2,3] systems, as input. The SLAM system is used to estimate camera poses and generate a 3D point cloud map of the structure, which forms the spatial database. Each map point in the point cloud stores the 3D pose in world coordinate $m^w$, its index, and a reference to the index of the source keyframe where it was first detected. Tagging the index of the 3D point to the index of the keyframe provides the spatial index of the database. Our system takes the point cloud map and aligns it to the 3D CAD model to update the geometry of the spatial database. This allows the user to perform a spatial query through the 3D CAD model by use of mouse clicks on the model to search for the desired images. Our system also provides a metric data analysis tool to analyze the queried image. Finally, a textured 3D model of the structure is generated to serve as an overall visual as-built documentation of the structure.

   Building inspection for quality assurance often involves analyzing images and detecting any anomaly or defect in the construction. For a multi-storied large building, it is necessary that the images are collected from a close range for proper visualization and inspection. Using a UAS allows us to collect high resolution building images from close range, and perform detailed image processing and analysis for anomaly detection. This was necessary as ground based images would not be able to provide the required fidelity to perform accurate building inspections.

The tools that have been developed to generate as-built documentation are:

   **Spatial Query for Images:** Our contribution is to allow a user to perform an efficient spatial query for images by aligning the database to the 3D CAD model. The user may easily visualize the 3D CAD model, making our implementation convenient for anyone to search for images of a specific region of interest without having to search through time intensive videos.

   **Metric Data Analysis:** Our system accurately calculates the metric distance between any two places of the as built structure. A 3D CAD model-image correspondence is used to ensure the robustness of the calculation. To facilitate a better experience, our system has a magnifier which works on mouse over to ensure a higher precision for selecting the desired pixel.

   **Visualization through 3D textured model:** The proposed system also generates a 3D textured model with high-resolution images to create a virtual reality of the scene. The virtual reality is visualized with an OpenGL window implemented with a moving camera to allow the user to move around the scene. The 3D textured model enables engineers or facility mangers to visually assess the building and track construction progress.

## 2 Related Research

There have been more than a million small UAS sold in the United States over the past few years according to news reports [4, 5]. As the use of small UAS grows the need for cost-effective methods for accessing and processing data will grow.

Contractors are already using small UAS to gather information about their worksite and inspect structures. In [6, 7] small UASs are used to construct a detailed 3D map of work sites. Others have used small UAS to inspect existing structures [7]. In [8] they outline the potential applications of UAS in the new construction such as monitoring the build process, creating "as-built" documentation and automated defect detection. Researchers are currently working on creating the algorithms needed to exploit this potential. One such system presented in [9] was used to aid in the creation of "as-built" documentation. Their work generated reasonable output, but the dimensions had more than 5% error.

The D4AR modeling [10] uses an unordered collection of images of the structure to generate the underlying geometric model by using a Structure-from-Motion (SfM). They solve the similarity transform between the model and 3D point cloud found from the SfM using minimum user inputs to transform the SfM coordinate system to the 3D CAD model's coordinate system to allow the aligning the SLAM photos to the CAD model. There are popular methods being used to create 3D mesh models by using images. Such methods have been used in [11, 12, 13, 14, and 15].

A different approach was taken by [16]. Instead of creating a 3D mesh model using images, they rely on an existing semantic 3D CAD model known as Building Information Model (BIM). This modeling is widely available nowadays to facilitate easier construction as it provides prior detailed information about the building or structure to be constructed.

Methods described in [10, 11, 12, 13, 14, 15] use images to create underlying geometry. Such methods are not useful for construction sites because it only provides visual information of an already built structure. Construction sites require a pre-designed CAD model or blueprints to be able to have ground truth for the construction and compare it to the as-built structure to detect anomalies. Considering this, our method uses a 3D CAD model designed with AutoCAD Revit as [16].

In this work, out design considerations are governed by the need of engineers on a construction site. PCL construction provided us with a checklist which specifically requested the need for an accurate metric data analysis as a priority. Another important tool required in that checklist was the need for a user-friendly tool to easily search for images through the 3D CAD model.

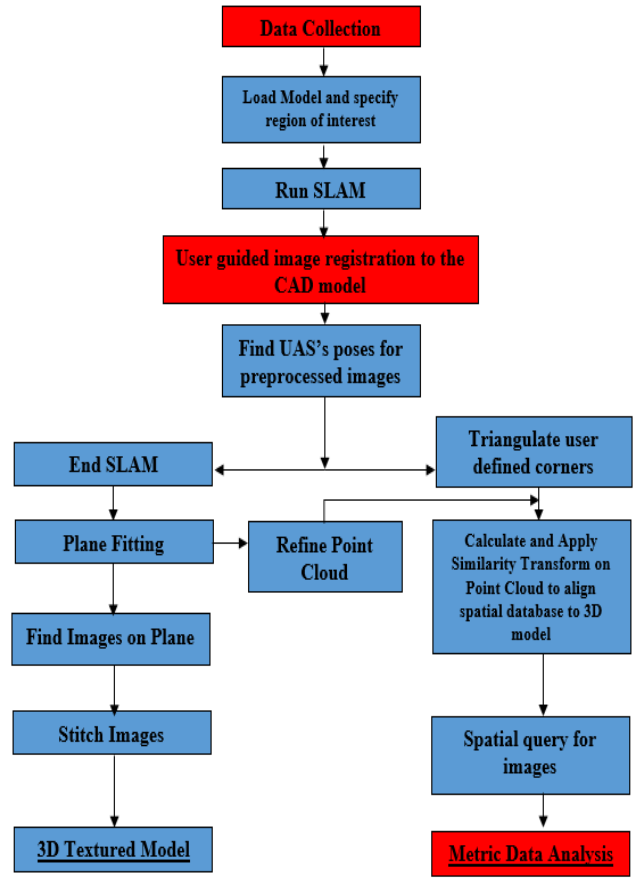

Figure 1. The process flow chart for system overview. Blue boxes indicate processes that require no user interaction. Red boxes indicate steps where user input is required.

## 3 System Overview

Figure 1 outlines our system. First, the user specifies the region of interest for inspection from the model (Section 5.1). The system leverages user input to register the first keyframe to the 3D CAD model. Our system prompts the user to specify a four-point correspondence between the model and the 1st keyframe at the beginning to aid the registration process (Section 5.1). This registration explores the pose of the UAS in model coordinates using the p3p algorithm (Section 5.1). The system takes a 3D point cloud map from the SLAM as input. The point cloud is refined with a plane-fitting algorithm (Section 5.3). A similarity transform is applied to the point cloud to transform the point cloud to the 3D CAD model coordinate system (Section 5.2) thus updating the geometry of the spatial database. The updated spatial database uses the spatial index to run a query for images of the desired region specified by the user (Section 5.4). The image returned from the spatial query may further be used to evaluate distance in the as-built structure. To generate the 3D textured model, our system retrieves images on the planes and stitches them together to texture map those images to the specified zone of the textured model on a different window (Sec 5.6). The metric information is mainly intended to find dimensions of various entities in the as-built structure (Section 5.5). The user must click on different corners of the model to inspect the size of that specific entity i.e. windows, doors, the length of a column etc. After clicking on the model, the corresponding image of that area will be seen, and the user will specify the desired dimension with mouse clicks.

## 4  Data Collection

Our system utilized the DJI Spreading Wings S1000+ as our UAS. One of the most important features of the S1000+ is a low gimbal mounting bracket which enables a wide range of possible viewing angles and camera motions [17]. It was fully compatible with the Zenmuse Z15 camera gimbals from DJI which stabilizes the camera to the desired orientation during flight. A Canon EOS 5D Mark III camera has been used to capture video of the scene. The camera has a 22.3 Megapixel CMOS sensor. High-Definition video with a resolution of 1920×1040 was collected at 30 frames per second.

The SLAM requires a calibrated camera system. Calibration images were taken before every flight. The 3D CAD model used was loaded in the system using Open Asset Import Library (Assimp) and OpenGL.

## 5  Methodology

Our system requires inputs from the user for registering images to the 3D CAD model. In this section, the methodology of the system has been described. The image registration process, alignment of 3D CAD model to the spatial database, and spatial query has been explained in detail.

With the registration of the images, plane fitting is performed which facilitates further processing such as 3D textured reconstruction, and metric information extraction.

### 5.1  Image Registration to 3D CAD Model

The user must provide four 3D-2D point correspondences between the 3D CAD model and the first keyframe. The Perspective-Three-Point problem determines the position and orientation of the camera while capturing the first keyframe in the model coordinate system. The p3p algorithm provides up to four solutions which are disambiguated by using a fourth point [18]. The user clicks on four corners of an entity, such as a window, from the model. That entity must be seen in the first keyframe so that the user can then click on corresponding four corners of the same entity from the image in a sequential order.

### 5.2  Alignment of Spatial Database to the 3D CAD Model

Initially the 3D point cloud, and the camera poses generated by the SLAM are expressed in the camera coordinate system with the first camera pose set as the identity. In order to align the spatial database to the 3D CAD model, the 3D point cloud along with the camera poses need to be transformed to the CAD model coordinate system from the camera coordinate system. Since monocular SLAM suffers from scale ambiguity, it is required to find the metric scale for the SLAM generated map that matches the scale of the 3D CAD model that has been provided.

Transformation between the CAD model coordinate system and the camera coordinate system is performed by first creating a set of 3D-3D points correspondences between the two coordinate systems, and then by computing the similarity transforms between the 3D point correspondences. That similarity transform is applied to the point cloud and camera poses to transform them to the 3D CAD model coordinate system.

To create the 3D-3D correspondences, the points selected by the user as described in section 5.1 were used. The selected points on the CAD model constitute the 3D points set in the CAD model coordinate system. Now we need their corresponding 3D points in the camera coordinate system. This is done by using the 2D image points selected by the user in the first keyframe. Since it is not guaranteed that the selected 2D points correspond to any 3D point in the SLAM generated map, their 3D positions are estimated by finding their 2D correspondences in the second keyframe. As our system takes an ordered photo collection, the specified entity should appear in the second keyframe too. Our system creates small 11×11 patches around the selected points of the first keyframe. Then a normalized cross-correlation based matcher finds the corresponding four 2D points in the second keyframe. Once a set of 2D-2D points correspondences are found, they are triangulated to estimate their 3D positions in the camera coordinate system.

Once the set of 3D-3D correspondences are found between the CAD model coordinate and camera coordinate systems, they are used to compute the similarity transform,

$$T = \begin{bmatrix} R & \vec{t} \\ \vec{0} & s \end{bmatrix} \quad (1)$$

between the CAD model and the camera coordinate system. The similarity transform is computed using our implementation of the Horn's method [19]. The similarity transform is then used to transform the point cloud and the camera poses to the CAD model coordinate system.

### 5.3 Plane Fitting

Assuming the structure to be piecewise planar, our system implemented a plane fitting algorithm for both assigning points to planes as well as reject outliers. A RANSAC [20] based plane fitting method was implemented that uses a voting scheme for assigning points in the 3D point cloud to individual planes. For a set of 3D data points $\{P_i(x_i, y_i, z_i); i = 1, \ldots, N\}$, where $N$ is total number of 3D points in the point cloud, the plane equation has been defined as,

$$ax + by + cz + d = 0 \quad (2)$$

where, $a, b, c$ are slope parameters and $d$ is the of the plane from the origin.

Planes are extracted by randomly constructing different planes from point cloud data. Three random points are sampled from the point cloud and checked for collinearity. If collinear or coincident, new points are considered. A plane hypothesis is created, and this process is repeated for a predefined number of times. The resulting candidate planes are scored against all points in the cloud to validate the candidate plane. In a candidate plane, the points that falls onto that plane, votes for the plane. The total vote for a candidate plane is the score of that plane. After a predefined number of trials, the candidate plane having the highest score is validated as a plane. Points voting for the valid plane are tagged to that plane and removed from the plane fitting consideration. The procedure is then repeated on the remainder of the point cloud to find subsequent planes. The planes are finally made robust using least-square constrains. All the points not included in any plane are considered as outliers and removed from the cloud. Figure 2(a) shows the region selection for inspection and 2(b) shows the point cloud generated with SLAM after being refined with plane fitting.

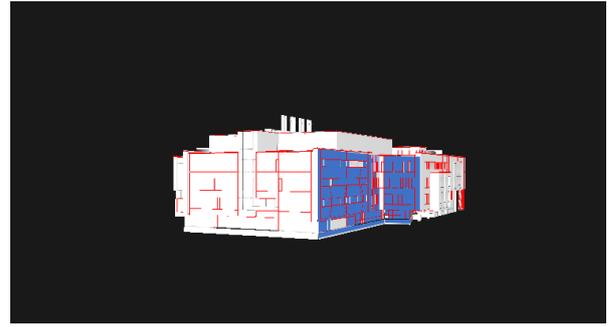

(a)

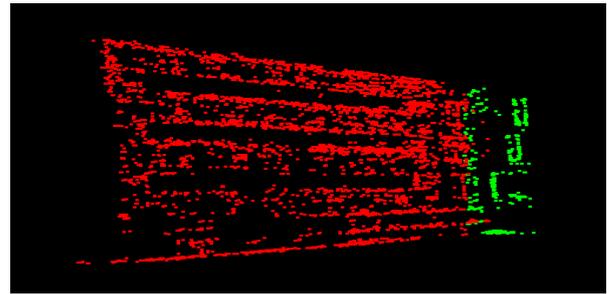

(b)

Figure 2. (a) The blue area indicates the region selected for inspection (b) Point cloud generated by SLAM and refined with plane fitting.

### 5.4 Spatial Query for Images

As the source keyframes are tagged along with each point in the cloud, a spatial index of images is used for efficient spatial image query. The entries in the spatial index depend on the vertices location in the model coordinate system.

The user must click on the area or entity on the visualization window for inspection. A ray casting method was used to select the 3D location of the mouse click in the model. The selected 3D point of the model was snapped to the nearest vertex, so that the user does not have to click precisely. The nearest point of the point cloud to the selected vertex is selected as the entry to spatial index for query and the associated source keyframe to that point is returned.

### 5.5 Metric Data Analysis

Our system assumes the user desires distance from one vertex to another to check the dimensions of different entities. By using the spatial query for an image, the user finds the image of the entity to be inspected. A binary image of the queried image is created and contours are found by using [21]. A Doughlas-Peucker [22] algorithm has been used to find rectangular shapes from the contours found. The contour area is calculated to make sure the area is larse enough to be considered as a window.

The actual height of the window is taken during alignment of the spatial database to the 3D CAD model. The pixel distance of the height of the window then sets a scale factor. The user clicks on two points of the image to find the 3D distance and the nearest windows scale factor is used to find the 3D distance. A magnifier has been implemented to aid the accuracy of clicking on the image. Figure 3 shows the detected windows by using this system.

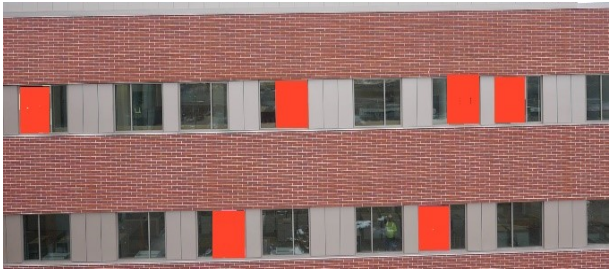

Figure 3. Automatically detected windows as described in section 5.5.

### 5.6 3D Textured Model Generation

All keyframes for each plane are then stored in a vector to be stitched together as [23]. Due to a high number of key-points our system takes in every tenth image from the keyframe vector and stitches them together. The user must specify a required area to be inspected during image registration to the 3D mesh model. That process takes in the boundaries for each plane in the specified region. That boundary is used to texture map the stitched image to its corresponding plane. A photorealistic 3D texture model is created in this manner.

### 6 Experimental Result

In this section, we compare our proposed algorithm against the latest version of the state of the art Pix4D mapper (version 3.1) [24]. The primary focus of this research is to incorporate a BIM to access the point cloud which provides us a prior knowledge on how the building was intended to be constructed. The CAD model provides entry to the spatial database eliminating the need to perform a dense reconstruction as Pix4D. We will show that not having a dense reconstruction does not affect our metric data calculations while and outperforming Pix4D.

The point cloud generated by the SLAM have been aligned to the 3D CAD model by a similarity transform. Each point has a keyframe associated with it, which creates the spatial index of images. The user clicks on the desired entity. The nearest point from the point cloud to the vertex is selected, working as an entry to the spatial index for query and efficiently searches for the spatial image data. Figure 4 shows some of the outputs using the spatial query.

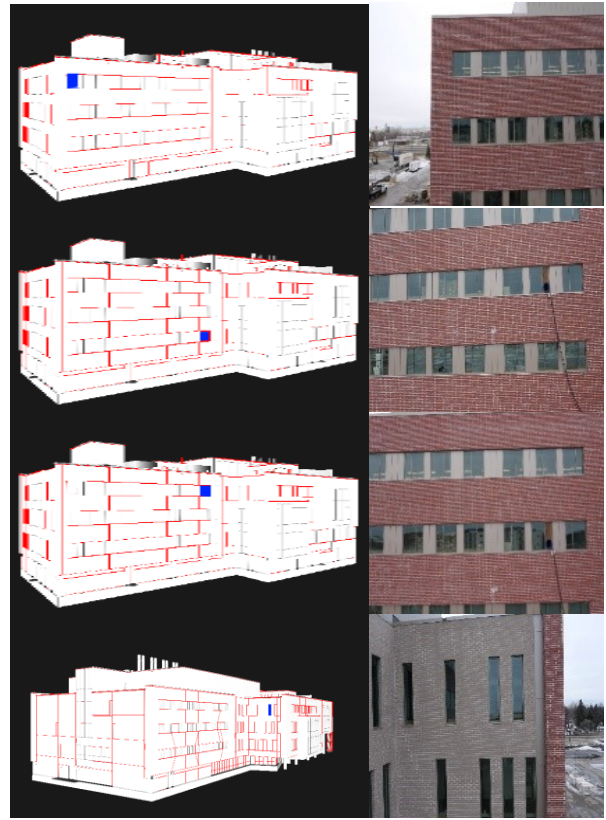

Figure 4. Spatial query: The window to be inspected in the model is marked with blue square and the image of that region found through the spatial query.

After a query the user may choose to check dimensions of various entities which is the metric data analysis in our system. We have compared our dimensional calculations to Pix4D's calculations. Our system showed significant improvements over Pix4D in Metric data extraction. However, Pix4D requires geotagged images to assign scale and orientation. It is not always possible to have geotag information with images, so our algorithm was delepoed keeping this in mind and geotags was not included in the proposed algorithm. As our system does not require geotagged images, we provided the scale manually to Pix4D. To apply a scale constraint into Pix4D the recommended process is to click on both vertices from the dense point cloud that Pix4D generates and provide the accurate metric distance. Moreover, it is recommended by Pix4D to correct the vertices in at least two of the corresponding images. Our system applies scale by using two consecutive images. As a direct comparison, we applied scale to the Pix4D model using the dense point cloud and two images for scale correction.

Pix4D does recommend more scale constraints as it will provide better accuracy. This is true for both Pix4D, as well as the proposed algorithm. Figure 5 below provides the process to apply scale using Pix4D mapper.

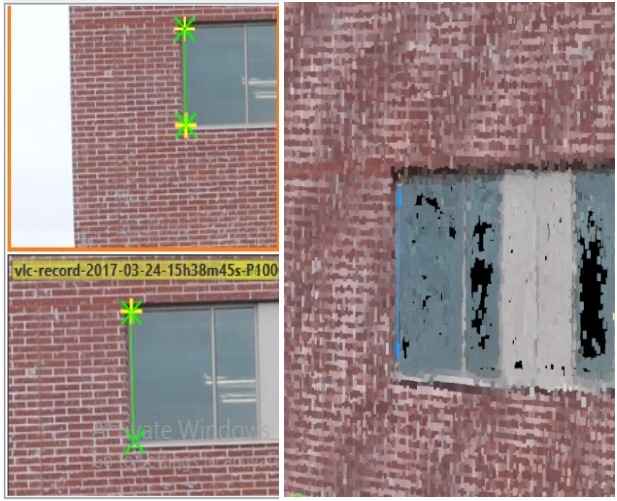

Figure 5. Applying scale constraint on Pix4D mapper.

After applying scale constraints, we have tested the height and width of 15 windows providing 60 total distances for each method. For verification, these same distances were manually measured. The actual width and height were 2.01168 meters(m) (6.6 feet) and 1.8288 meters(m) (6 feet). The CAD model provides these distances as 1.8288 meters for both width and height. The scale was applied based on the height of the windows, as provided by the CAD model, on both the proposed model and Pix4D. We will show that the proposed system and Pix4D both determine the building was not constructed per the CAD model dimensions.

The proposed model resulted in a mean squared error (MSE) of 31.9 cm$^2$ whereas Pix4D mapper's MSE was of 45.6 cm$^2$. For Pix4D's width calculation, it has a standard deviation of 4.92 cm, as opposed to the proposed system's standard deviation of 4.28 cm. For height calculation, Pix4D's standard deviation resulted in 4.17 cm, where the proposed algorithm provided a standard deviation of 3.27 cm. Pix4D had a combined standard deviation of 6.45 cm where the proposed algorithms combined standard deviation was 5.39 cm. Figure 6 provides the width and height calculations along with the actual.

A t-test was performed on the width and height averages to verify the statistical significance of the calculated errors. Tables 1 and 2 show the resulting t-tests from the two-sample width and height data assuming unequal variances. In both cases, an ∝ of 0.01 was used. This shows we are 99% confident that the proposed method is statistically different from Pix4D. The null hypotheses for both cases were set as there are no significant difference.

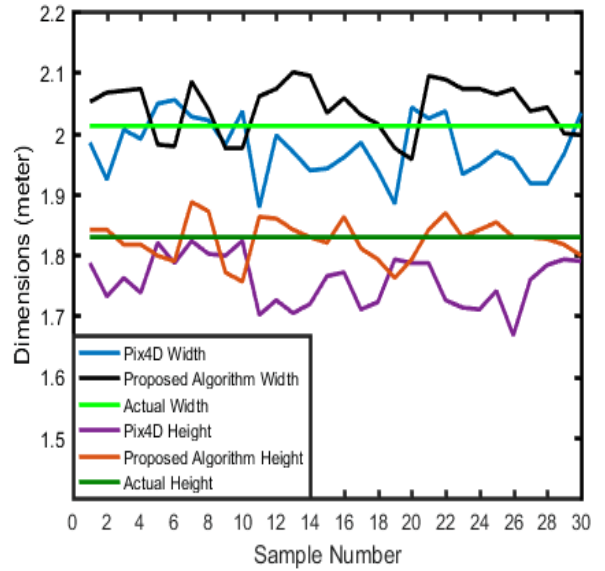

Figure 6. Dimensions calculated with Pix4D and Proposed Algorithm along with theit actual values.

Table 1. t-test results on two sample widths assuming unequal variances

|  | Pix4D Width | Proposed Algorithm Width |
|---|---|---|
| Mean | 1.97673 | 2.04084 |
| t Stat | -5.38338 | |
| t Critical 2 tail | 2.66487 | |
| P two-tail | 1.4E-06 | |

Table 2. t-test results on two sample heights assuming unequal variances

|  | Pix4D Height | Proposed Algorithm Height |
|---|---|---|
| Mean | 1.75758 | 1.82494 |
| t Stat | -6.95956 | |
| t Critical 2 tail | 2.66822 | |
| P two-tail | 4.4E-09 | |

Due to the $p_{value}$ for both resulted in a much lower value than the ∝ value, the null hypothesis can be rejected. We can therefore state that the mean average in the proposed algorithm were significantly closer to the actual values than Pix4D.

An analysis of variance (ANOVA) was performed on the variances of the data analyses to verify the statistical significance of the sample variables. Table 3 provides the ANOVA output.

Table 3. Two-factor ANOVA with replication results on both width and height with $\alpha = 0.01$

| Source of Variation | F | P-value | F Critical |
|---|---|---|---|
| Sample | 73.39466 | 5.25E-14 | 6.858521 |
| Width/Height | 803.691 | 5.6E-54 | 6.858521 |
| Interaction | 0.044844 | 0.832588 | 6.858521 |

Pix4D and the proposed algorithm are represented as the two samples. The $p_{value}(sample)$ was lower than $\alpha$. The $F_{Crirtical}(sample)$ is lower than $F(sample)$ showing both samples have significant differences. The second row in Table 3 represents the effect of both samples on width and height calculations and the $p_{value}(width/height)$ is much less than $\alpha$. Therefore, we can reject the null hypothesis and conclude both systems are significantly different on the width and height calculations. The $p_{value}(interaction)$ calculated a value greater than $\alpha$, showing we can conclude the interaction between width and height have no significant difference, meaning the effect of width or height does not depend on one another.

We can therefore assume the proposed algorithm provides significantly more accurate results than Pix4D data analysis.

Another tool developed was for visualization through 3D textured model. To visualize the 3D texture model, the OpenGL library was used with a moving camera implementation. This allows the user to roam through the 3D textured model to look for any visual anomalies. We have compared the proposed system to Pix4D mapper's textured reconstruction. For a fair comparison, we have selected the high resolution texture mapping option for Pix4D. Figure 7 shows some of the images taken from the proposed systems 3D textured modelling.

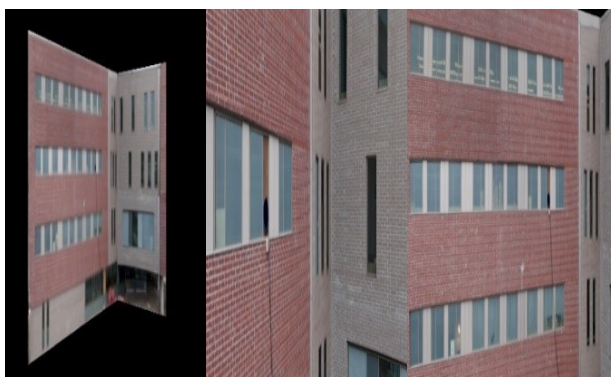

Figure 7. 3D texture model generated with the proposed system from various camera locations

Figure 8 shows the differences between the proposed system and Pix4D's 3D texture models.

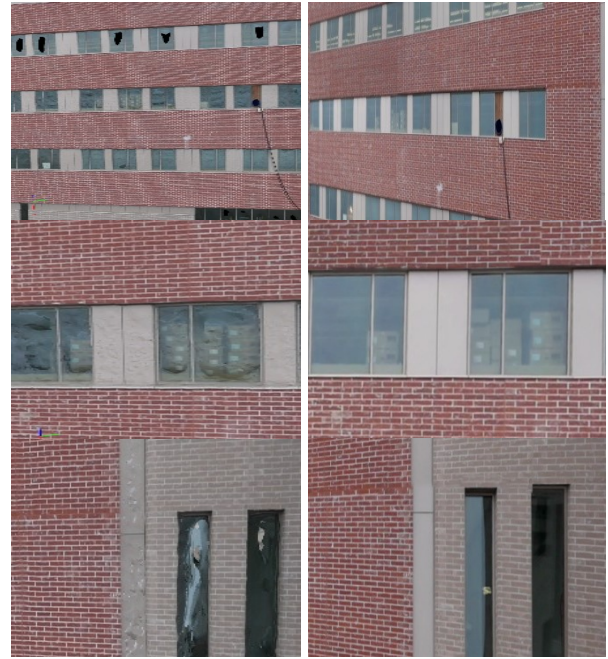

Figure 8. Left Column: Pix4D's textured model
Right Column: Proposed Algorithm's textured model

Pix4D's textured reconstruction results in clearly visible holes and artifacts. Windows are not realistic and provide visible anomalies. Straight lines are distorted resulting in wavy lines, unusable for detailed visual inspections. The proposed method does not have any of these artifacts resulting in a photorealistic rendition.

## 7 Conclusion

In this system, we have successfully demonstrated spatial query into the spatial database through the provided 3D CAD model. We have performed metric data analysis along with visualization through a 3D textured model. A comparison with the state of the art Pix4D mapper was given where our proposed system has been proven to significantly improve the metric data analysis and provided better and photorealistic 3D textured model. Our system is easy to use and does not require the user to have any previous knowledge of visualization, rendering or CAD software. . Interesting steps towards further research could be window detection with deep convolutional neural networks, automatic scheduling monitoring and temporal navigation using the 3D CAD model, and optimizing the system to implement on an on-board environment or mobile devices.